\documentclass[]{arxiv_preprint}

\usepackage{amsmath}
\usepackage{amsfonts}
\usepackage{amssymb}
\usepackage{colortbl}
\usepackage{flafter}
\usepackage{wrapfig}
\usepackage{xspace}

\providecommand{\method}{Robust-WAM\xspace}

\title{\textcolor{metablue}{Robust-WAM}: Bridging Generative Pretraining and Semantic Foresight in World-Action Models}

\author[1,\ast,\dagger]{Haodong Yan}
\author[1,\ast]{Junfeng Li}
\author[1,\ast]{Junjie He}
\author[1]{Zhide Zhong}
\author[2]{MingMing Yu}
\author[1]{Wenxuan Song}
\author[1]{Jiaguan Zhu}
\author[1]{Yangyang Zheng}
\author[1]{Yuqiao Du}
\author[1]{Jiadi You}
\author[3]{Yingjie CAI}
\author[3]{Xu Yan}
\author[3]{Guanyi Zhao}
\author[3]{Bingbing Liu}
\author[1,\ddagger]{Haoang Li}

\affiliation[1]{The Hong Kong University of Science and Technology (Guangzhou), Guangzhou, China}
\affiliation[2]{Beihang University, Beijing, China}
\affiliation[3]{Huawei Foundation Model Department}
\contribution[\ast]{Equal contribution}
\contribution[\dagger]{Project Leader}
\contribution[\ddagger]{Corresponding author.}

\abstract{%
  Mainstream World-Action Models (WAMs) adapt pretrained video generation models
(VGMs) for robot control, transferring their learned dynamics prior for action
prediction. These VGMs are typically trained in a variational autoencoder (VAE)
latent space. However, the VAE latent space is optimized for pixel reconstruction,
which rewards fine appearance detail and leaves the action prediction fragile under
visual shifts. Recent works build WAMs in semantic latent space, which are more
robust to appearance shifts. However, these models cannot leverage the large-scale
VGM pretraining that exists only in VAE space. To overcome this dilemma, we propose
Robust-WAM, a general post-training method for video-generation-based WAMs that
preserves the VAE-based generative path and adds a lightweight semantic foresight
alignment objective on the action stream. This retains the large-scale VGM
pretraining while grounding actions in appearance-invariant dynamics that stay
reliable under illumination shifts and other visual out-of-distribution conditions.
Specifically, we employ learnable query tokens to bring future-scene semantics into
the action stream by aligning their output hidden states with the semantic foresight
of future ground-truth frames. To establish the temporal correspondence between each
query and the future step it describes, we give it the positional encoding of the
matching action tokens. Experiments on out-of-distribution generalization simulation
benchmarks and a real-robot setup show that our Robust-WAM consistently improves the
success rates of multiple WAM baselines without sacrificing in-distribution
performance.

}

\date{August 2026}
\metadata[Project Page]{\url{https://haodong-yan.github.io/robust-wam-project-page/}}

\begin{document}

\maketitle

\providecommand{\method}{Robust-WAM}

\section{Introduction}

World-Action Models (WAMs) have recently emerged as a powerful paradigm for robot manipulation~\citep{wu2024gr1,cheang2024gr2,guo2024pad,bi2025motus,li2026causalwam,kim2026cosmospolicy,ye2026dreamzero,yuan2026fastwam,zhong2026dualcotvla,zhong2025flowvla,song2025reconvla,yan2026svam,yan2025scar}.
They harness the dynamics priors of pretrained video generation models (VGMs) to ground action generation. These priors, learned from web-scale video, capture how the world evolves: how objects move, make contact, and interact over time~\citep{wan2025wan,nvidia2025cosmos}.

\begin{wrapfigure}{r}{0.485\textwidth}
\centering
\includegraphics[width=\linewidth]{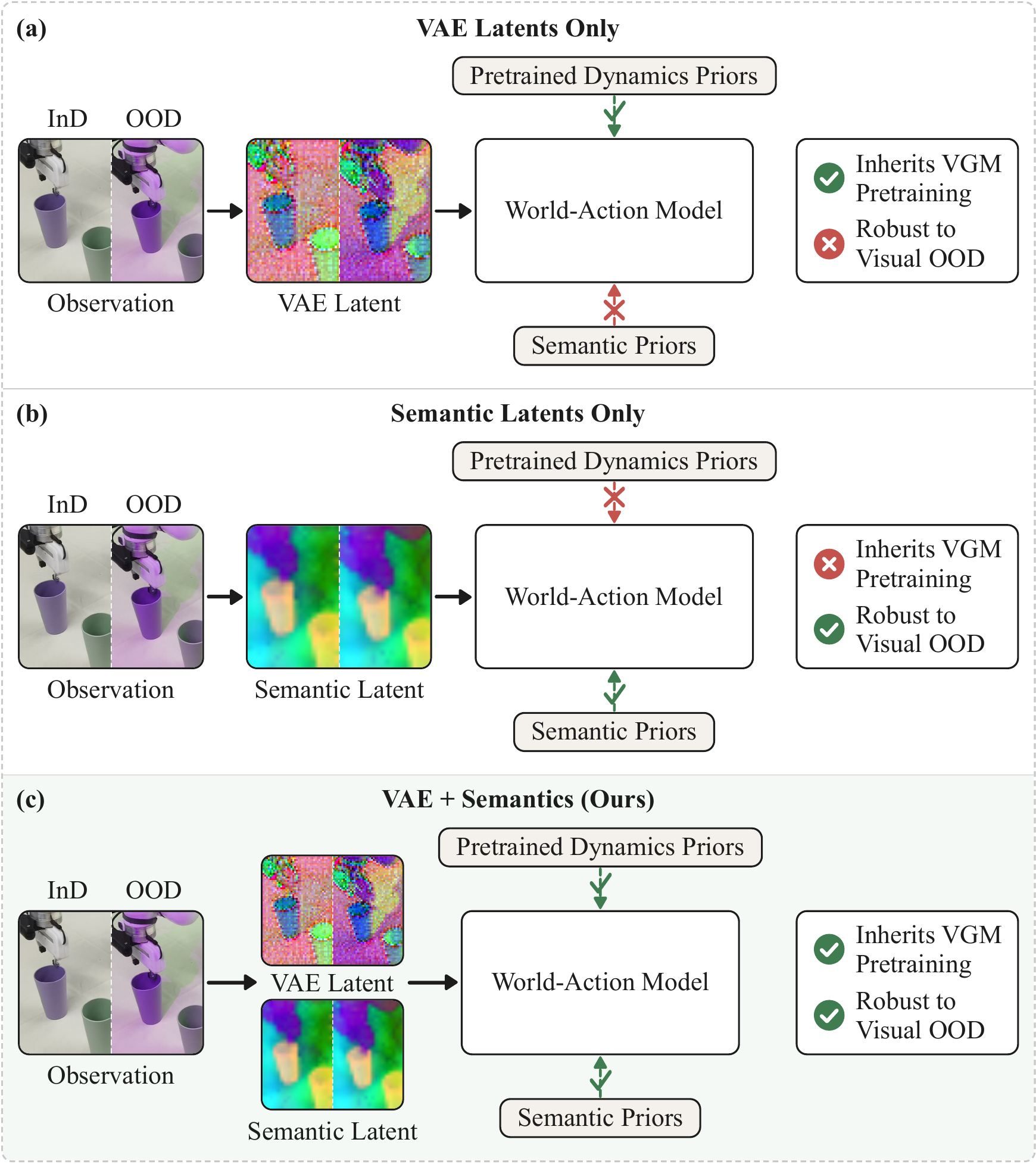}
\caption{
    (a) VAE-latent WAMs inherit the dynamics priors of large-scale VGM pretraining, but their latents move with appearance and the policy breaks under visual shifts.
    (b) Semantic-latent world models are robust to visual shifts, but cannot leverage the pretrained dynamics priors.
    (c) Our \method{} keeps the VAE generative path and injects semantic priors into the action stream, obtaining both.}
\label{fig:teaser}
\end{wrapfigure}

Although the dynamics priors from large-scale pretrained VGMs are a great help to action generation, they are learned in a variational-autoencoder (VAE) latent space trained for pixel reconstruction.
This objective forces the space to preserve realistic appearance, such as texture and illumination. A large share of the priors is therefore devoted to appearance detail that is irrelevant to actions.
Inheriting this appearance bias, the action stream is easily disrupted under visual out-of-distribution conditions, such as photometric or hue changes~\citep{fei2025liberoplus,zhang2026wamrobustness}.
To overcome this limitation, recent methods~\citep{zhou2024dinowm,lda1b2026,chen2026lawam} rebuild visual generation in a semantic latent space, such as DINO~\citep{simeoni2025dinov3} or V-JEPA~\citep{assran2025vjepa2}.
These models are visually robust, but they abandon the strong dynamics priors of large-scale pretrained VGMs; regaining that knowledge requires costly re-pretraining on large-scale video data~\citep{assran2025vjepa2}.
\enlargethispage{2\baselineskip}
This leaves WAMs facing a fundamental trade-off (illustrated in Figure~\ref{fig:teaser}): VAE-based WAMs inherit the strong dynamics priors of large-scale VGM pretraining but remain fragile to appearance changes, whereas semantic-latent WAMs are robust to appearance changes but cannot leverage that large-scale pretraining.

\newpage

\begin{figure*}[t]
\centering
\includegraphics[width=\textwidth]{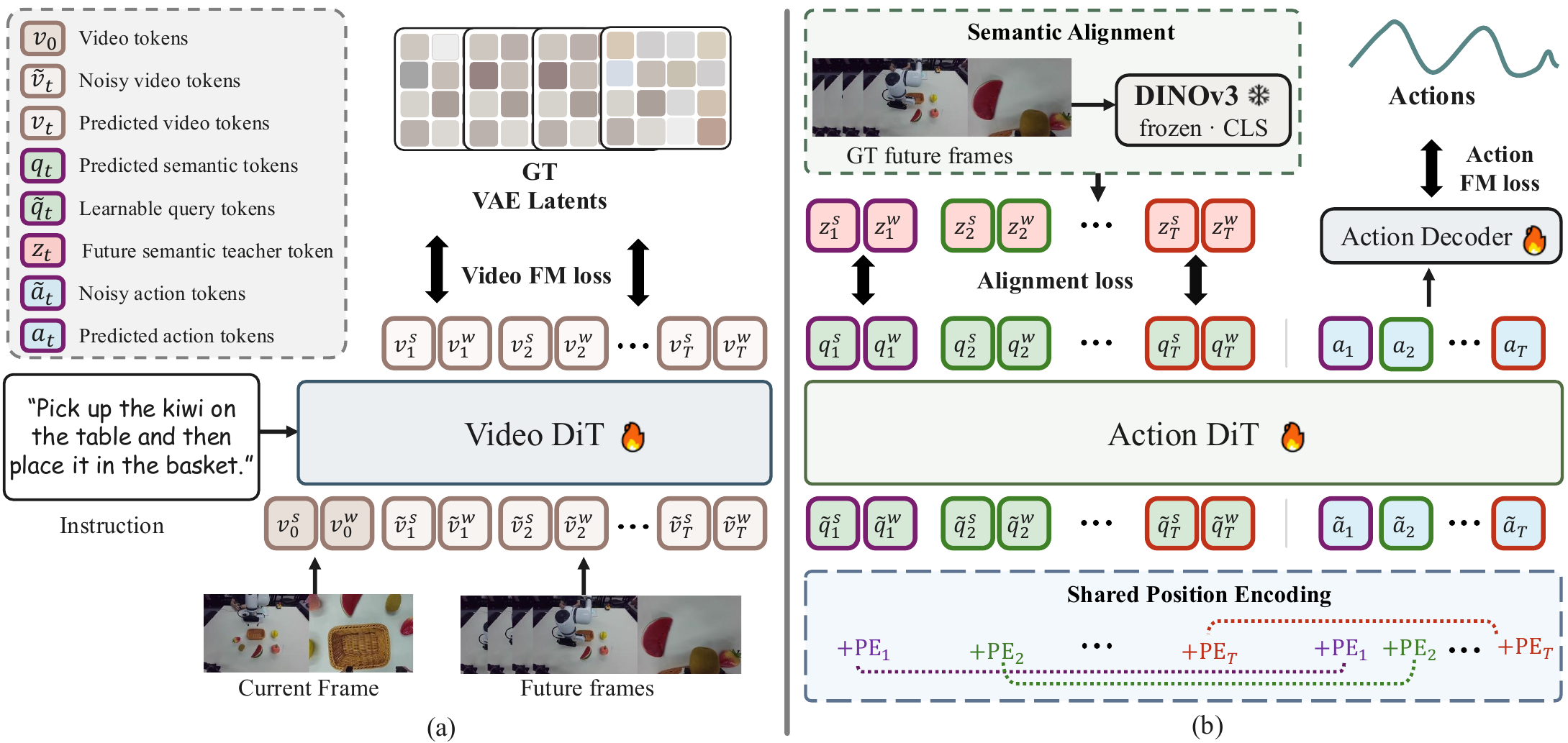}
\caption{Overview of \method{}.
(a) We inherit the video branch of the original WAM as it is.
The Video DiT keeps denoising the VAE latents of the future frames under its own flow-matching loss, so the \textbf{large-scale VAE-space pretraining} is retained.
(b) The upper part is the \textbf{semantic alignment}.
For each future step and camera view we prepend a learnable query token to the noised action tokens, and after the Action DiT processes the whole sequence we align the output of each query with the frozen DINOv3 CLS embedding of the corresponding ground-truth future frame, which is an appearance-invariant summary of the scene content.
The lower part is the \textbf{shared positional encoding (PE)}. Each query reuses the positional encoding of the action tokens at the step it describes, which tells the model how far into the future that query is looking.
The teacher and the alignment head are only needed during training, so at inference they are dropped and only the queries stay in the sequence.}
\label{fig:method}
\end{figure*}

To address this dilemma, we introduce \method{}, a general post-training method for video-generation-based WAMs that preserves the VAE-based generative path and adds a lightweight semantic foresight alignment objective on the action stream.
Rather than replacing the whole latent space, it corrects the appearance bias directly in the action representation.
Specifically, as shown in Figure~\ref{fig:method}, we prepend learnable query tokens to the noised action tokens of the action Diffusion Transformer (DiT) stream. Then we align the output hidden states of these query tokens with the frozen DINOv3~\citep{simeoni2025dinov3} CLS embedding of the corresponding ground-truth future frame. We employ the DINOv3 CLS embedding as the alignment target because self-supervised semantic features are largely invariant to appearance: the CLS token summarizes scene content, such as objects and their spatial arrangement, while discarding the texture and illumination details.
We align to future rather than current frames so that the queries encode where the demonstrated actions lead, providing the action stream with appearance-invariant foresight.
To help the model relate these queries to the future steps they represent, each query carries a temporal positional encoding taken from the WAM's own encoding at the corresponding action step.

We evaluate \method{} on two out-of-distribution (OOD) generalization benchmarks whose test conditions deliberately depart from training.
On RoboTwin clean$\rightarrow$random~\citep{chen2025robotwin2}, policies are trained only on clean demonstrations with fixed initializations and evaluated on a randomized split with unseen object poses, textures, and lighting; on LIBERO-Plus, policies trained on the standard LIBERO suites~\citep{liu2023libero} face thousands of perturbed task variants spanning camera, lighting, background, layout, and other axes.
Across both benchmarks and multiple WAMs, including LingBot-VA~\citep{li2026causalwam}, GE-Act~\citep{liao2025genie}, and FastWAM~\citep{yuan2026fastwam}, \method{} consistently improves success rates over the corresponding baselines without sacrificing in-distribution (InD) performance, indicating that semantic foresight alignment restores OOD generalization without sacrificing the benefits of large-scale VAE-space pretraining.

In summary, our contributions are threefold:
\begin{itemize}
    \item We propose \method{}, a general post-training method that bridges
    video-generation WAMs with semantic foresight without discarding the dynamics
    priors acquired through large-scale video pretraining, thereby improving
    robustness to visual OOD shifts.
    
    \item To instantiate this bridge, we introduce a semantic foresight alignment objective where learnable query tokens are aligned with the frozen DINOv3 CLS embeddings of future frames, giving the action stream appearance-invariant foresight. Each query reuses the positional encoding of its corresponding action tokens, establishing its temporal correspondence to the future step it describes.
    \item Experiments across multiple WAMs on two OOD generalization benchmarks (RoboTwin clean$\rightarrow$random and LIBERO-Plus) and a real-robot setup show that \method{} consistently improves OOD success rates without sacrificing InD performance.
\end{itemize}

\section{Related Work}

\paragraph{World-Action Models.}
Most World-Action Models (WAMs) couple pretrained VGMs with action generation, so that
robot manipulation can benefit from the dynamics priors learned from web-scale
video~\citep{shen2026wamsurvey}. Early works generate a visual future and recover the
actions that realize it with a separate inverse-dynamics
model~\citep{du2023unipi,pai2025mimicvideo}, while more recent works predict video and
actions in a unified or tightly coupled model, letting action prediction directly
consume generative features~\citep{wu2024gr1,cheang2024gr2,guo2024pad,bi2025motus,
li2026causalwam,kim2026cosmospolicy,ye2026dreamzero,yuan2026fastwam}. They differ
mainly in the space in which the future is modeled.
The mainstream line works in the VAE latent space in which the pretrained
VGM was trained~\citep{wan2025wan,nvidia2025cosmos}. Optimized for pixel
reconstruction, this space ties the inherited prior to appearance, leaving the action
stream sensitive to visual shifts that leave the task
unchanged~\citep{fei2025liberoplus,zhang2026wamrobustness}.

A second line instead predicts the future directly in a semantic feature space,
following the original view of world models as prediction in an abstract representation
rather than raw pixels~\citep{ha2018world,lecun2022path}. LDA-1B first scales this into a
semantic-latent \emph{world-action} model that jointly denoises future DINO states and
action chunks~\citep{lda1b2026}, and later work extends this semantic-latent
world-action modeling~\citep{chen2026lawam,wang2026repwam}. Such semantic latents are consistently found
to favor control over the reconstruction latents optimized for visual
fidelity~\citep{balestriero2024reconstruction,nilaksh2026reconstruction}. These models
are robust but give up the web-scale VGM the VAE-space WAMs were built to reuse: a
predictor for the new space must be trained from scratch~\citep{assran2025vjepa2}.
\method{} bridges the two, leaving an existing WAM's VAE tokenization and denoising
path untouched and adding semantic supervision only into the action stream, where the
appearance bias actually harms control.

\paragraph{Representation Alignment in Robot Learning.}
Representation alignment~\cite{yu2024representation} regularizes a model's internal features toward a frozen
pretrained target, rather than rebuilding the latent space. In robot learning, such methods differ mainly in \emph{what} they align to. One line targets
\emph{spatial} structure, aligning a policy's features to the geometric embeddings
of a frozen 3D foundation model so that a 2D backbone gains spatial awareness at no
inference cost~\citep{li2025spatial,lin2025evo}. A second line
targets \emph{foresight}: FLARE adds learnable future tokens to an action denoiser
and aligns them with future-observation embeddings~\citep{zheng2025flare}, FRAPPE
extends this to multiple visual foundation models~\citep{zhao2026frappe}, and
FutureVLA distills separately pretrained visuomotor embeddings into downstream
vision-language-action (VLA) policies~\citep{xu2026futurevla}.
Our \method{} differs on two points. First, these methods perform future
representation alignment in standalone policies rather than within a pretrained
video-generation WAM, whereas \method{} injects semantic queries into a WAM that
already carries a large-scale \emph{dynamics prior}. Second, they do not explicitly
couple future queries with the corresponding action steps. We introduce a shared
temporal PE mechanism that provides more precise temporal grounding for semantic
foresight during action generation.

\section{Method}
\label{sec:method}

\subsection{Preliminaries: World-Action Models}
\label{sec:method_prelim}
A world-action model (WAM) couples a pretrained VGM with an action DiT.
We use $\theta$ to denote the trainable WAM parameters and $d$ to denote the hidden width of the action stream. 
Given the current observation $o_t$ and a language instruction $\ell$, it jointly
denoises two streams: the VAE latents
$\mathbf{x} = \mathcal{E}(o_{t+\Delta}, o_{t+2\Delta}, \dots, o_{t+T_f\Delta})$ of
$T_f$ future frames, where $\mathcal{E}$ is the frozen encoder of the VGM's
pretrained VAE and successive frames are $\Delta$ control steps apart, and the
action chunk $\mathbf{a} = a_{t:t+H-1}$ executed over the next $H$ control steps, where $a_t$ denotes the action at control step $t$.
The chunk enters the action DiT as $L$ action tokens
$\mathbf{h}^a \in \mathbb{R}^{L \times d}$, one per control step, so that $L = H$
and token $i$ carries the action $a_{t+i-1}$.
Both streams are trained with a flow matching objective.
For a flow time $s \sim \mathcal{U}[0,1]$ and independent Gaussian noise $\epsilon, \epsilon' \sim \mathcal{N}(\mathbf{0}, \mathbf{I})$, each stream is corrupted by linear interpolation between data and noise, $\mathbf{x}^{s} = s\,\epsilon + (1-s)\,\mathbf{x}$ and $\mathbf{a}^{s} = s\,\epsilon' + (1-s)\,\mathbf{a}$, so that $s{=}0$ leaves the sample clean and $s{=}1$ is pure noise.
The video DiT $v^{v}_{\theta}$ regresses the velocity that transports the noised latents back to the clean ones,
\begin{equation}
  \mathcal{L}_{\text{video}} = \mathbb{E}\Big[\big\|
    v^{v}_{\theta}(\mathbf{x}^{s}, s \mid o_t, \ell) - (\epsilon - \mathbf{x})
  \big\|^2\Big],
\end{equation}
The two streams are not denoised independently: the action blocks read the video
stream through attention. We write
$\mathbf{H}^{v} = \mathcal{H}_{\theta}(\mathbf{x}^{s}, s \mid o_t, \ell)$ for the
video hidden states that the action blocks attend to. In a unified video--action
transformer this is the video segment of the joint sequence, and in a
mixture-of-transformers (MoT) WAM with a separate action expert it is the
per-layer states of the video tower. The action DiT $v^{a}_{\theta}$ then regresses
the action velocity while attending to $\mathbf{H}^{v}$,
\begin{equation}
  \mathcal{L}_{\text{act}} = \mathbb{E}\Big[\big\|
    v^{a}_{\theta}(\mathbf{a}^{s}, s \mid \mathbf{H}^{v}, o_t, \ell)
    - (\epsilon' - \mathbf{a})
  \big\|^2\Big],
\end{equation}
where both expectations are taken over the training data, the flow time, and the Gaussian noise, and $\|\cdot\|_2$ denotes the Euclidean norm over all elements of the corresponding prediction.
The two streams are optimized together,
\begin{equation}
  \mathcal{L}_{\text{WAM}} = \mathcal{L}_{\text{video}}
    + \lambda_{a}\, \mathcal{L}_{\text{act}},
\end{equation}
where $\lambda_a$ is the action-loss weight specified by the original WAM training recipe.
Action prediction therefore inherits the dynamics priors of the pretrained VGM
through $\mathbf{H}^{v}$.
At inference the action chunk is produced by integrating $v^{a}_{\theta}$ from $s{=}1$ to $s{=}0$ starting from a Gaussian sample.

\subsection{Semantic Foresight Alignment}
\label{sec:method_alignment}
To improve the robustness of WAMs while preserving their pretrained
video-generation formulation, we add a representation alignment objective to
the action stream.
We prepend $K$ learnable query tokens to the noised action tokens and align their
output hidden states with semantic features of the ground-truth future frames.
The action stream thus acquires appearance-invariant foresight, while the WAM's own
VAE tokenization, video-generation objective, and video--action architecture
remain unchanged.

\paragraph{Per-Frame Semantic Query Tokens.}
We introduce $K$ learnable query tokens
$\mathbf{q} \in \mathbb{R}^{K \times d}$, one for each future time step and
camera view, and prepend them to the action sequence:
$\mathbf{h} = [\mathbf{q}; \mathbf{h}^a] \in \mathbb{R}^{(K+L) \times d}$.
The WAM already equips its action stream with a positional encoding, which we
write as a map $\Phi(\cdot, p)$ applied to a token at position index $p$. We reuse
this same map for the queries. We index the queries by $k = (j-1)C + c$ for future
frame $j \in \{1, \dots, T_f\}$ and camera view $c \in \{1, \dots, C\}$. Frame $j$
is the observation $o_{t+j\Delta}$ reached after executing $a_{t+j\Delta-1}$, which
is action token $j\Delta$, so the action step matching frame $j$ is
\begin{equation}
  \tau_j = \min\big(j\Delta,\; L\big) \in \{1, \dots, L\},
  \label{eq:tau}
\end{equation}
where the clamp applies only when the video horizon $T_f\Delta$ outruns the action
horizon $H$, in which case the trailing frames share the final action step. The
action blocks then receive the position-encoded sequence
$\hat{\mathbf{h}} \in \mathbb{R}^{(K+L) \times d}$,
\begin{equation}
  \hat{\mathbf{h}}_n =
  \begin{cases}
    \Phi\big(\mathbf{q}_n,\, \tau_{\lceil n / C \rceil}\big), & \text{if } n \le K, \\[3pt]
    \Phi\big(\mathbf{h}^a_{n-K},\, n-K\big), & \text{if } n > K,
  \end{cases}
\end{equation}
for $n = 1, \dots, K+L$. Each query thus takes the positional code of the action
step it describes, and the action tokens keep their original indices
$1, \dots, L$. The augmented sequence is then processed by the unmodified action
blocks.

\paragraph{Semantic Foresight Targets.}
For each training sample, we extract the CLS embeddings of a frozen DINOv3
encoder~\citep{simeoni2025dinov3} from the $T_f$ ground-truth future frames of
each of the $C$ camera views, yielding targets
$\mathbf{z}^{\ast} \in \mathbb{R}^{K \times d_z}$, where
$K = T_f \times C$. A CLS embedding summarizes scene content while remaining
comparatively insensitive to texture and illumination, so each target states in
appearance-invariant terms where the demonstrated action leads. Their number and
dimension are given in Sec.~\ref{sec:exp_setup}.

\paragraph{Alignment Objective.}
The $N$ action blocks map $\hat{\mathbf{h}}$ to an output sequence
$\bar{\mathbf{h}} \in \mathbb{R}^{(K+L) \times d}$, which we split back into query
and action outputs. A linear head $g: \mathbb{R}^{d} \to \mathbb{R}^{d_z}$ maps the
query outputs $\bar{\mathbf{h}}_{1:K}$ to the target space, while the original
action head receives exactly its usual $L$ action tokens $\bar{\mathbf{h}}_{K+1:K+L}$.
We optimize a per-query cosine alignment loss,
\begin{equation}
  \mathcal{L}_{\text{align}}
  = \frac{1}{K} \sum_{k=1}^{K}
  \left(1 - \cos\!\big(g(\bar{\mathbf{h}}_k),\; \mathbf{z}^{\ast}_k\big)\right),
  \label{eq:align}
\end{equation}
and add it to the WAM objective,
\begin{equation}
  \mathcal{L} = \mathcal{L}_{\text{WAM}}
    + \lambda_{\text{align}} \, \mathcal{L}_{\text{align}},
\end{equation}
where $\cos(\cdot,\cdot)$ denotes cosine similarity and $\lambda_{\text{align}}$ balances the auxiliary alignment loss against the original WAM objective.
Eq.~\ref{eq:align} makes each query output describe the scene at its future step in appearance-invariant representation space.

The queries sit in the same sequence as the action tokens, so the action blocks
attend between the two.
Each query reads the action plan and the video stream through the WAM's existing
attention, and each action token reads the future semantics the queries carry.
Action generation is thus conditioned on features that ignore texture and
illumination, and stays stable when only appearance changes.
The teacher features are never provided as inputs to the queries during
training. They only supervise the query outputs through
$\mathcal{L}_{\text{align}}$. At inference, we therefore drop the DINOv3
teacher and the head $g$ while retaining the $K$ query tokens, so the query
inputs and the action-stream attention path remain unchanged.

\subsection{Integration with Different Architectures}
\label{sec:method_instantiation}

\paragraph{Action-Expert WAMs.}
For mixture-of-transformers (MoT) WAMs with a dedicated action expert, such as
GE-Act~\citep{liao2025genie}, FastWAM~\citep{yuan2026fastwam}, and
Motus~\citep{bi2025motus}, the queries are prepended directly to the expert's
token sequence. They join the expert's self-attention and cross-attend to the
video tower in the same way as the action tokens.

\paragraph{Unified WAMs.}
For a unified video-action WAM such as
LingBot-VA~\citep{li2026causalwam}, the queries are inserted into the action
segment and assigned the same sequence, frame, and noise IDs as the
corresponding action tokens. Under the original flex-attention mask rule, the
queries inherit the attention pattern of the action tokens, allowing them to
interact bidirectionally with action tokens at the same future step while
preserving the autoregressive ordering across steps.

\section{Experiments}
\label{sec:experiments}

We evaluate \method{} on simulation and real-robot manipulation to answer the following questions:
\begin{itemize}
    \item (Q1) Does semantic foresight alignment improve OOD robustness without sacrificing in-distribution performance?
    \item (Q2) Is \method{} broadly applicable across WAM architectures?
    \item (Q3) Which design choices make the alignment work?
    \item (Q4) Does \method{} improve robustness under real-world visual shifts?
\end{itemize}

\subsection{Experimental Setup}
\label{sec:exp_setup}

\paragraph{Implementation Details.}
Each WAM jointly predicts $T_f{=}8$ future frames and the action chunk. We keep each
WAM's own action horizon $H$ and frame stride, so the query-to-action-step
map $\tau_j$ of Eq.~\ref{eq:tau} is fixed by the released recipe rather than tuned.
For alignment
we attach one query token per (future frame, camera) and match it to the DINOv3 CLS of
that frame, giving $K{=}16$ for LIBERO and the real-robot setup, each with one
third-person and one wrist camera, and $K{=}24$ for RoboTwin, with one
third-person and two wrist cameras. The target is the frozen
DINOv3~\citep{simeoni2025dinov3} ViT-B/16 CLS ($d_z{=}768$), extracted offline once,
with alignment weight $\lambda_{\text{align}}{=}0.1$. At inference the DINOv3
teacher and the alignment head are dropped, leaving only the $K$ query tokens in the
action stream.
Within each WAM, the baseline and the \method{} variant share the same training
data, schedule, and optimization settings. Across WAMs, the training regimes
differ, as each follows its own released recipe and is evaluated on the benchmark
reported by its authors.

\paragraph{Methods for Comparison.}
We compare against two families. \emph{VLA methods} map observations directly
to actions without modeling the future: OpenVLA~\citep{kim2024openvla},
WorldVLA~\citep{cen2025worldvla},
NORA~\citep{hung2025norasmallopensourcedgeneralist},
UniVLA~\citep{bu2025univla}, $\pi_0$~\citep{black2024pi_0},
$\pi_0$-FAST~\citep{pertsch2025fast},
OpenVLA-OFT~\citep{kim2025fine}, StarVLA~\citep{ye2026starvla}, and
GR00T-N1.7~\citep{bjorck2025gr00t}.
\emph{WAM methods} predict the future to guide action and can be distinguished
by their prediction space: video-generation WAMs operate in the VAE latent
space (FastWAM~\citep{yuan2026fastwam}, GE-Act~\citep{liao2025genie}, and
LingBot-VA~\citep{li2026causalwam}), whereas LDA-1B~\citep{lda1b2026}
models future dynamics in a semantic latent space.

\subsection{OOD Robustness on LIBERO-Plus (Q1)}
\label{sec:exp_q1}

\paragraph{Setup.}
We apply \method{} to two action-expert WAMs, FastWAM~\citep{yuan2026fastwam} and GE-Act~\citep{liao2025genie}. 
All policies are trained on the four standard LIBERO suites~\cite{liu2023libero} and evaluated on the four clean LIBERO suites and on all perturbed tasks of LIBERO-Plus~\cite{fei2025liberoplus}, spanning seven perturbation axes, namely camera viewpoint, lighting, background texture, object layout, robot initial state, language, and sensor noise.

\newpage
\paragraph{Results.}
\begin{wraptable}{r}{0.49\textwidth}
\centering
\caption{
\textbf{Comparison on Clean LIBERO}~\cite{liu2023libero} \textbf{and
LIBERO-Plus}~\cite{fei2025liberoplus}. Average success rate (\%) over all
evaluation trials. \textbf{Bold} indicates the best result.
}
\label{tab:q1_main}
\begingroup
\scriptsize
\setlength{\tabcolsep}{3pt}
\begin{tabular*}{\linewidth}{@{\extracolsep{\fill}}lcc@{}}
\toprule
Method & LIBERO & LIBERO-Plus \\
\midrule
\multicolumn{3}{@{}l}{\textbf{VLAs}} \\
OpenVLA~\citep{kim2024openvla}     & 76.5 & 15.6 \\
WorldVLA~\citep{cen2025worldvla}   & 81.8 & 25.0 \\
NORA~\citep{hung2025norasmallopensourcedgeneralist} & 87.9 & 39.0 \\
UniVLA~\citep{bu2025univla}        & 95.2 & 43.9 \\
$\pi_0$~\citep{black2024pi_0}      & 94.2 & 53.6 \\
$\pi_0$-FAST~\citep{pertsch2025fast} & 85.5 & 61.6 \\
OpenVLA-OFT~\citep{kim2025fine}    & 97.1 & 69.6 \\
\midrule
\multicolumn{3}{@{}l}{\textbf{WAMs}} \\
LDA-1B~\citep{lda1b2026}           & 90.6 & 45.5 \\
FastWAM~\citep{yuan2026fastwam}    & 97.6 & 49.7 \\
\rowcolor{gray!15}
\quad + \method{}                  & \textbf{97.9}\,($+0.3$) & 58.9\,($+9.2$) \\
GE-Act~\citep{liao2025genie}       & 96.5 & 78.0 \\
\rowcolor{gray!15}
\quad + \method{}                  & 97.3\,($+0.8$) & \textbf{80.9}\,($+2.9$) \\
\bottomrule
\end{tabular*}
\endgroup
\end{wraptable}

As shown in Table~\ref{tab:q1_main}, GE-Act with \method{} achieves the best
LIBERO-Plus performance among all compared methods, reaching $80.9\%$. It
outperforms the strongest baseline, GE-Act, by $2.9$ points.
The consistent gains on both FastWAM and GE-Act demonstrate the broad
applicability of \method{} across different action-expert WAMs. Specifically,
\method{} improves FastWAM from $49.7\%$ to $58.9\%$ on LIBERO-Plus, a gain of
$9.2$ points, and improves GE-Act from $78.0\%$ to $80.9\%$, a gain of $2.9$
points. Meanwhile, clean LIBERO performance increases from $97.6\%$ to
$97.9\%$ on FastWAM and from $96.5\%$ to $97.3\%$ on GE-Act. These results
show that the improvements in OOD robustness do not come at the cost of
in-distribution performance.
The comparison with LDA-1B further highlights the importance of large-scale
video-generation pretraining. LDA-1B, a semantic-latent WAM without such
pretraining, reaches $90.6\%$ on clean LIBERO and $45.5\%$ on LIBERO-Plus,
whereas GE-Act with \method{} achieves $97.3\%$ and $80.9\%$, respectively.
Together with the controlled improvements over FastWAM and GE-Act, this
contrast suggests that semantic representations are more effective when used
to enhance, rather than replace, the dynamics priors of a pretrained
video-generation WAM.

\begin{wraptable}{r}{0.49\textwidth}
\centering
\caption{
\textbf{Per-axis results on LIBERO-Plus}~\citep{fei2025liberoplus}.
Success rate (\%) over all evaluation trials within each perturbation axis.
}
\label{tab:q1_axes_main}
\begingroup
\scriptsize
\setlength{\tabcolsep}{1.6pt}
\begin{tabular*}{\linewidth}{@{\extracolsep{\fill}}lcccc@{}}
\toprule
Axis & FastWAM & \shortstack{FastWAM\\+ ours} & GE-Act &
\shortstack{GE-Act\\+ ours} \\
\midrule
Layout   & 60.5 & 67.8\,($+7.3$)  & 82.8 & 85.4\,($+2.6$) \\
Camera   & 16.3 & 28.7\,($+12.4$) & 51.9 & 54.8\,($+2.9$) \\
Init     & 44.4 & 43.4\,($-1.0$)  & 77.9 & 82.3\,($+4.4$) \\
Language & 68.1 & 73.6\,($+5.5$)  & 79.4 & 82.9\,($+3.5$) \\
Light    & 78.9 & 90.2\,($+11.3$) & 95.4 & 95.2\,($-0.2$) \\
BG       & 52.3 & 62.6\,($+10.3$) & 84.7 & 87.1\,($+2.4$) \\
Noise    & 37.9 & 56.8\,($+18.9$) & 81.6 & 85.1\,($+3.5$) \\
\midrule
Overall  & 49.7 & 58.9\,($+9.2$) & 78.0 & 80.9\,($+2.9$) \\
\bottomrule
\end{tabular*}
\endgroup
\end{wraptable}

Table~\ref{tab:q1_axes_main} further breaks down the improvements by
perturbation axis. On FastWAM, the largest gains occur under sensor noise
($+18.9$), camera perturbations ($+12.4$), lighting changes ($+11.3$), and
background changes ($+10.3$), while robot initial state is the only axis that
does not improve. On the already strong GE-Act baseline, \method{} improves
six of the seven perturbation axes, led by robot initial state ($+4.4$) and
language ($+3.5$). The gains also extend beyond photometric shifts. Layout
perturbations alter the spatial arrangement of the scene, while camera
perturbations change the viewpoint and image projection. Improvements under
both settings suggest that the semantic target also benefits robustness to
geometric and viewpoint variations.

\subsection{Generalization Across Different WAM Architectures (Q2)}
\label{sec:exp_q2}
\paragraph{Setup.}
To evaluate whether our post-training method generalizes across WAM
architectures, we apply \method{} to LingBot-VA~\citep{li2026causalwam}, a
Wan2.2-based~\citep{wan2025wan} unified autoregressive WAM that interleaves
video and action tokens in a shared flex-attention sequence. We train the
LingBot-VA baseline and its \method{} variant on the same $2{,}500$ bimanual
demonstrations from the official clean split of
RoboTwin~\citep{chen2025robotwin2} for $15$K steps. Both models are evaluated
on the clean split and a randomized split with unseen object poses, textures,
and lighting, using $100$ episodes per task. The randomized split therefore provides a
demanding test of policy generalization and robustness. As described in
Sec.~\ref{sec:method_instantiation}, adapting \method{} to LingBot-VA only
requires inserting the queries into the action segment and assigning them the
same sequence, frame, and noise IDs as the corresponding action tokens.

\newpage
\paragraph{Results.}
\begin{wraptable}{r}{0.49\textwidth}
\centering
\caption{
\textbf{Comparison on RoboTwin clean$\rightarrow$random}~\citep{chen2025robotwin2}.
We report the mean task success rate (\%) across all evaluated tasks under
the clean and random settings. \textbf{Bold} indicates the best result in
each column.
}
\label{tab:q2_robotwin}
\begingroup
\scriptsize
\setlength{\tabcolsep}{4pt}
\begin{tabular*}{\linewidth}{@{\extracolsep{\fill}}lcc@{}}
\toprule
Method & clean & random \\
\midrule
\multicolumn{3}{@{}l}{\textbf{VLAs}} \\
StarVLA~\citep{ye2026starvla}      & 58.1 & 10.6 \\
GR00T-N1.7~\citep{bjorck2025gr00t} & 43.6 & 20.7 \\
\midrule
\multicolumn{3}{@{}l}{\textbf{WAMs}} \\
LDA-1B~\citep{lda1b2026}           & 53.0 & 12.3 \\
LingBot-VA~\citep{li2026causalwam} & \textbf{81.2} & 29.8 \\
\rowcolor{gray!15}
\quad + \method{}                  & \textbf{81.2} & \textbf{34.4} \\
\bottomrule
\end{tabular*}
\endgroup
\end{wraptable}

As shown in Table~\ref{tab:q2_robotwin}, LingBot-VA with \method{} achieves the
highest randomized success rate among all compared methods, reaching $34.4\%$.
Compared with the corresponding LingBot-VA baseline, \method{} improves the
randomized success rate from $29.8\%$ to $34.4\%$, a gain of $4.6$ points,
while preserving its $81.2\%$ clean success rate. This shows that \method{} is a
general post-training method: the same approach that improves the action-expert
WAMs in Q1 also transfers to a fundamentally different, unified video-action
architecture.
Consistent with the LIBERO-Plus results in Q1, LDA-1B, which lacks large-scale
video-generation pretraining, reaches only $53.0\%$ on the clean split and $12.3\%$
under randomization, far below LingBot-VA with \method{} ($81.2\%$ and $34.4\%$).
This again shows that semantic representations are best used to enhance, not
replace, the dynamics priors of a pretrained video-generation WAM.

\subsection{Ablation Study (Q3)}
\label{sec:exp_q3}

\paragraph{Setup.}
We ablate two design choices of \method{} on the full LIBERO-Plus suite, using
FastWAM with DINOv3 CLS targets and temporally indexed future queries as the
full model. First, the alignment-target variants compare DINOv3 CLS, DINOv3
Patch, DepthAnything3 (DA3)~\citep{lin2025depthanything3}, and the combined DINOv3 CLS~+~DA3 targets. For DINOv3 Patch, an MLP in
the alignment head compresses the teacher patch sequence to $K$ query targets,
while DINOv3 CLS~+~DA3 uses separate query groups for the two targets. Second, the
query-design variants compare the full temporal query design with variants
that remove temporal positional encoding or replace future-frame targets with
the current-frame target. All variants share the same data and training
schedule. 
The single-target and query-design variants use $K{=}16$ queries, whereas
DINOv3 CLS~+~DA3 uses two target-specific groups of $16$ queries, totaling
$32$ queries.

\paragraph{Results.}
\begin{wraptable}{r}{0.49\textwidth}
\centering
\caption{
\textbf{Ablations on LIBERO-Plus}.
We report average success rate (\%) over all evaluation trials.
\textbf{Bold} indicates the best result.
}
\label{tab:q3_ablation}
\begingroup
\scriptsize
\setlength{\tabcolsep}{3pt}
\begin{tabular*}{\linewidth}{@{\extracolsep{\fill}}lrr@{}}
\toprule
Variant & Overall & Improvement \\
\midrule
Baseline & 49.7 & --- \\
\midrule
\multicolumn{3}{@{}l}{\textbf{(i) Alignment target}} \\
DINOv3 Patch                   & 57.0 & +7.3 \\
DA3                            & 56.8 & +7.1 \\
DINOv3 CLS + DA3               & 56.5 & +6.8 \\
\rowcolor{gray!15}
DINOv3 CLS (ours)              & \textbf{58.9} & \textbf{+9.2} \\
\midrule
\multicolumn{3}{@{}l}{\textbf{(ii) Query design}} \\
w/o temporal PE                & 55.0 & +5.3 \\
Current-frame target ($o_t$)   & 52.7 & +3.0 \\
\rowcolor{gray!15}
Full temporal query (ours)     & \textbf{58.9} & \textbf{+9.2} \\
\bottomrule
\end{tabular*}
\endgroup
\end{wraptable}

Table~\ref{tab:q3_ablation} shows that both the alignment target and temporal
query design affect robustness. For alignment targets, DINOv3
CLS~\citep{simeoni2025dinov3} improves the baseline by $+9.2$ points and
achieves the strongest overall performance. Depth Anything~3
patch features~\citep{lin2025depthanything3} improve the baseline by $+7.1$
points. DINOv3 patch tokens also provide a substantial $+7.3$-point gain, but
remain below the compact CLS target. The DINOv3 CLS~+~DA3 variant reaches
$56.5\%$, below both DINOv3 CLS ($58.9\%$) and DA3 ($56.8\%$), suggesting that
simply combining semantic and geometric targets does not provide complementary
gains and that the added complexity of jointly optimizing two query groups and
alignment targets may hinder optimization. We therefore adopt DINOv3 CLS, which
achieves the strongest overall result using a single target vector per future
frame and view. For query design, the full temporal query design reaches
$58.9\%$, outperforming the variant without temporal positional encoding
($55.0\%$) by $3.9$ points and the current-frame variant ($52.7\%$) by $6.2$
points. These comparisons demonstrate the importance of both future-step
indexing and future semantic targets.

\begin{figure*}[!t]
\centering
\includegraphics[width=\textwidth]{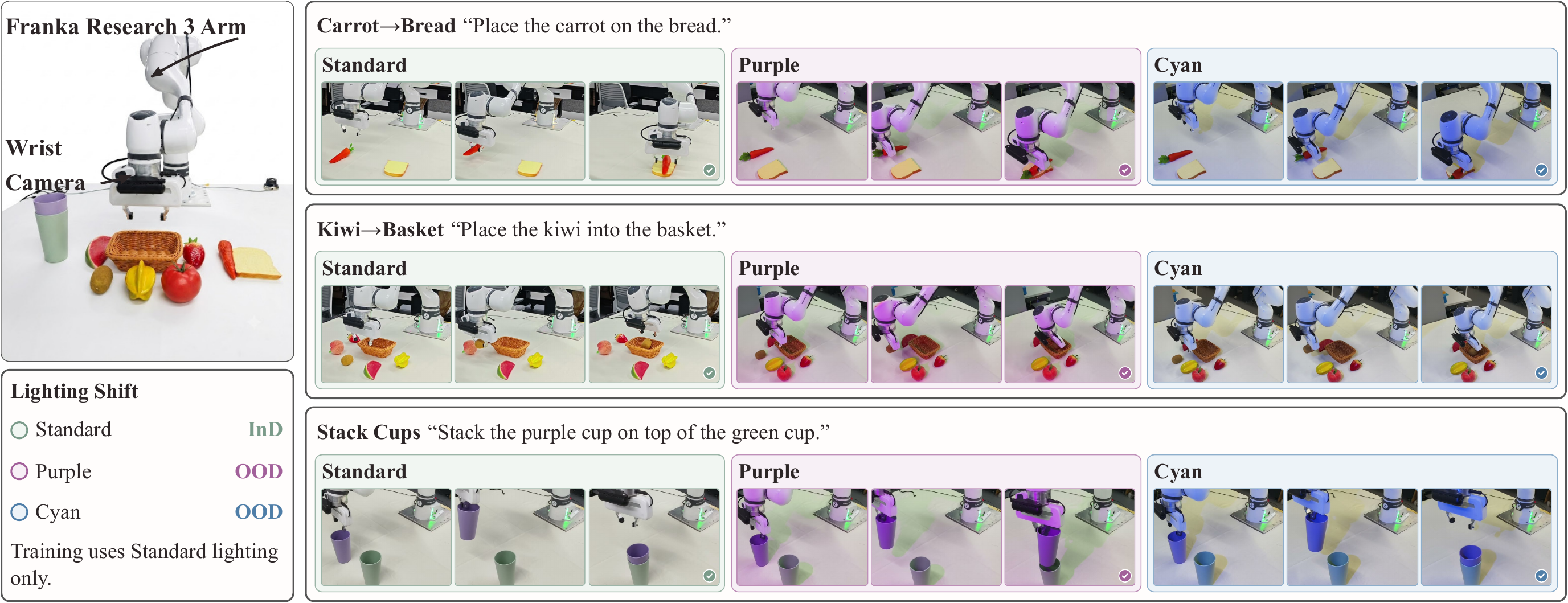}
\caption{
\textbf{Real-world tasks under illumination shifts.} Three tabletop
manipulation tasks on a Franka Research 3 arm observed by a fixed third-person
camera and a wrist camera. We train under standard illumination and evaluate
under two unseen illumination conditions, purple and cyan.
}
\label{fig:real_world}
\end{figure*}

\begin{table*}[!t]
\centering
\caption{
\textbf{Real-world Evaluation.}
We report success rate (\%) over $25$ rollouts per task per condition.
\textbf{Bold} indicates the best result.
}
\label{tab:real_world}
\begingroup
\small
\setlength{\tabcolsep}{3pt}
\begin{tabular*}{\textwidth}{@{\extracolsep{\fill}}lcccccccccccc@{}}
\toprule
& \multicolumn{3}{c}{Carrot$\to$Bread} & \multicolumn{3}{c}{Kiwi$\to$Basket}
& \multicolumn{3}{c}{Stack Cups} & \multicolumn{3}{c}{Avg.} \\
\cmidrule(lr){2-4}\cmidrule(lr){5-7}\cmidrule(lr){8-10}\cmidrule(lr){11-13}
Method & In-D. & OOD & $\Delta_{\text{gap}}\downarrow$
       & In-D. & OOD & $\Delta_{\text{gap}}\downarrow$
       & In-D. & OOD & $\Delta_{\text{gap}}\downarrow$
       & In-D. & OOD & $\Delta_{\text{gap}}\downarrow$ \\
\midrule
FastWAM~\citep{yuan2026fastwam} & 76 & 40 & 36 & 52 & 12 & 40 & 64 & 28 & 36 & 64.0 & 26.7 & 37.3 \\
GE-Act~\citep{liao2025genie}    & 88 & 64 & 24 & \textbf{72} & 48 & 24 & 80 & 60 & 20 & 80.0 & 57.3 & 22.7 \\
\midrule
\method{} (GE-Act)              & \textbf{92} & \textbf{92} & \textbf{0} & \textbf{72} & \textbf{68} & \textbf{4} & \textbf{84} & \textbf{80} & \textbf{4} & \textbf{82.7} & \textbf{80.0} & \textbf{2.7} \\
\bottomrule
\end{tabular*}
\endgroup
\end{table*}

\subsection{Real-World Deployment (Q4)}
\label{sec:exp_q4}

\paragraph{Setup.}
To evaluate robustness under real-world visual shifts, we conduct experiments
on a Franka Research 3 platform across the three tabletop tasks shown in
Figure~\ref{fig:real_world}: Carrot$\to$Bread, Kiwi$\to$Basket, and Stack Cups.
Demonstrations are collected under a single standard illumination.
At evaluation time we keep the scene, the objects, and the camera fixed and change only the illumination, replacing the standard white light with a purple and then a cyan light.
Neither colored illumination appears anywhere in the demonstrations.
For each task and policy, we conduct $25$ rollouts under standard illumination
and $25$ under each unseen illumination, giving $50$ OOD rollouts in total.
We compare two WAM baselines, FastWAM and GE-Act, against \method{} built on the
stronger GE-Act backbone, with all policies trained on the same demonstrations
and training schedule.

\paragraph{Results.}
As shown in Table~\ref{tab:real_world}, both baselines degrade sharply under
the unseen illuminations: FastWAM drops from $64.0\%$ to $26.7\%$ and GE-Act,
the stronger of the two, from $80.0\%$ to $57.3\%$. Applied to GE-Act,
\method{} consistently improves performance across all three tasks, raising the
average OOD success rate from $57.3\%$ to $80.0\%$ and reducing the
in-distribution-to-OOD gap from $22.7$ to $2.7$ points, while preserving
in-distribution performance ($82.7\%$ versus $80.0\%$). The improvements are
consistent across individual tasks, ranging from $20$ to $28$ points of OOD
success rate. Since the scene geometry, object placement,
and instruction remain unchanged while only illumination varies, these
results suggest that future semantic alignment reduces sensitivity to
appearance shifts while preserving in-distribution performance.

\section{Conclusion}
In this paper, we present \method{}, a general post-training method, which resolves the dilemma between the strong dynamics priors of VAE-based WAMs and their fragility under visual shifts, showing that appearance robustness can be gained without giving up large-scale generative pretraining.
The recipe is simple and plug-and-play: for each future time step and camera view, a set of learnable query tokens is prepended to the action stream with a temporal positional encoding. The output of these query tokens is aligned to the frozen DINOv3 CLS embedding of the ground-truth future frame.
Experiments across two OOD benchmarks, RoboTwin clean$\rightarrow$random and LIBERO-Plus, and across multiple WAMs demonstrate that our \method{} consistently improves success rates on OOD situations without sacrificing in-distribution performance.
Our ablation studies further validate the effectiveness of two key design choices: using DINOv3 CLS features as the alignment target achieves the strongest overall performance with low computational overhead, while temporally indexed queries outperform current-frame and non-temporal alternatives.

\end{document}